\documentclass[10pt,twocolumn,letterpaper]{article}

\usepackage{times}
\usepackage{epsfig}
\usepackage{graphicx}
\usepackage{amsmath}
\usepackage{amssymb}

\usepackage{color}
\usepackage{multirow}
\usepackage[left=1in,top=1in,right=1in,bottom=1in,letterpaper]{geometry}
\usepackage{setspace}
\usepackage{wrapfig}
\usepackage{enumitem}
\usepackage{float}
\usepackage{nonfloat}

\def\M{\mathcal{M}}

\def\F{\mathcal{F}}

\def\T{\mathcal{T}}

\def\RR{\mathbb{R}}
\def\PT{\mathcal{P}}
\def\vb{\vec{b}}
\def\vn{\vec{n}}

\newcommand\bigzero{\makebox(0,0){\text{\huge0}}}

\usepackage[pagebackref=true,breaklinks=true,letterpaper=true,colorlinks,bookmarks=false]{hyperref}

\begin{document}
\date{}
\title{Parallel Transport Convolution: A New Tool for Convolutional Neural Networks on Manifolds}

\author{Stephan Schonsheck\\
Rensselaer Polytechnic Institute\\
Troy, NY\\
{\tt\small schons@rpi.edu}
\and
Bin Dong\\
Peking Unversity\\
Beijing, China\\
{\tt\small dongbin@math.pku.edu.cn}
\and
Rongjie Lai\\
Rensselaer Polytechnic Institute\\
Troy, NY\\
{\tt\small lair@rpi.edu}
}

\maketitle

\begin{abstract}
Convolution has played a prominent role in various applications in science and engineering for many years. It is the most important operation in convolutional neural networks. There has been a recent growth of interests of research in generalizing convolutions on curved domains such as manifolds and graphs. However, existing approaches cannot preserve all the desirable properties of Euclidean convolutions, namely compactly supported filters, directionality, transferability across different manifolds. In this paper we develop a new generalization of the convolution operation, referred to as parallel transport convolution (PTC), on Riemannian manifolds and their discrete counterparts. PTC is designed based on the parallel transportation which is able to translate information along a manifold and to intrinsically preserve directionality. PTC allows for the construction of compactly supported filters and is also robust to manifold deformations. This enables us to preform wavelet-like operations and to define deep convolutional neural networks on curved domains.
\end{abstract}

\section{Introduction}

Convolution is a fundamental mathematical operation that arises in many applications in science and engineering~\cite{Daubechies:1992ten,mallat2008wavelet}. Its ability to effectively extract local features, as well as its ease of use, has made it the cornerstone of many important techniques such as numerical partial differential equations and wavelets. More recently, convolution plays a fundamentally important role in convolutional neural networks (CNN)~\cite{lecun1990handwritten} which have made remarkable progress and significantly advanced the state-of-the-art in image processing, analysis and recognition~\cite{lecun1990handwritten,bengio2015deep,krizhevsky2012imagenet,ciresan2012deep,hinton2012deep,sermanet2013overfeat,leung2014deep,sutskever2014sequence}.

In Euclidean space $\RR^n$, the convolution of a function $f$ with a kernel (or filter) $k$ is defined as:
\begin{equation}
(f * k)(x) := \int_{\RR^n} k(x-y)f(y) dy.
\end{equation}
Unlike signals or images whose domain is shift invariant, functions defined on curved domains such as curved manifolds do not have shift-invariance. Therefore, to properly define convolutions on these curved domains, one of the key challenges is to properly define the translation operation. This is one of the main obstacles of generalizing CNN to manifolds.

There has been a recent surge of research in designing CNNs on manifolds or graphs. We refer the interested readers to \cite{bronstein2017geometric} for a review of recent progress in this area. These approaches can be classified into three categories: spectral patch based and group action methods. Spectral methods are based on projecting a signal onto the eigen (Fourier) space and using the convolution theorem to define convolution. Patch based methods use a patch operator to interpolate local geodesic discs on a certain given template. Group action based methods are defined on homogeneous space with a transitive group action.   Here, we briefly review some of these approaches.

Spectral methods defines convolution based on the convolution theorem.The convolution theorem states that, for any two functions $f$ and $g$: $\F (f * g) = \F(f)\cdot \F(g)$
where $\F$ is the Fourier transform and $\cdot$ denotes point wise multiplication. This theorem can be naturally generalized to functions on manifolds if we let $\F$ to be the projection operator onto the Laplace-Beltrami (LB) eigensystem. This method has proven effective to handle functions on a fixed domain, and can be applied to graphs as well \cite{hammond2011wavelets,bruna2013spectral,dong2015sparse,henaff2015deep}. However, these methods have two fundamental limitations. First, the uncertainly principle states that a function can have compact support in either the time or frequency domain, but not both. These methods normally use only a finite number of eigenfunctions in the Fourier domain. As a result the kernels that arise from this methods are not localized (i.e. not compactly supported in the spatial domain). The second major drawback to these methods is that since they rely on the eigensytem of the domain, any deformation of the domain will change the egensystem which in turn changes the filters. The high frequency LB eigenfunctions of a manifold are extremely sensitive to even small deformations. This means that anything designed for, or learned on, one manifold can only be applied to problems on the same domain. This limit the transferability of the spectral based methods.

Patch based methods are originally proposed in \cite{masci2015geodesic}. In this work the authors propose the use of a local patch operator to interpolate local geodesic discs of the manifold to a fixed template and develop a Geodesic Convolutional Neural Network (GCNN). Then for each point on the manifold, the convolution is calculated as the multiplication between the values of the kernel and the extracted patch on the template. To do so, they create a local polar coordinate system at each point. One drawback to this approach is that there is no natural way to choose the origin for the polar coordinate. To overcome this, the authors consider an angular pooling operation which evaluates all rotations of their kernel at each point and selects the orientation which maximizes the convolution in a point-wise fashion. Since the angular pooling operation is computed independently at each point, the selected orientation does not reflect the geometric structure of the base manifold and may not be consistent even for nearby points.
More recently, \cite{boscaini2016learning} proposes an anisotropic convolutional neural network (ACNN) by replacing the the aforementioned patch operator with an operator based on anisotropic heat kernels with the direction of isotropy fixed on the principle curvature at each point. Although this introduces a new hyper-parameter (the level of isotropy), it allows the kernels to be directionally aware. However, 
filters developed for applications on one manifold can only be applied to manifolds in which the local directions of principle curvature are the same. 

Group action based methods are recently discussed in~\cite{kondor2018generalization,cohen2018spherical,chakraborty2018h}. A typical application of these methods is to extend CNN on the unit sphere~\cite{cohen2018spherical}, where convolutional operations can be defined by transferring kernels on the unit sphere through the rotation group. This idea can be generalized to a manifold $\M$ with a transitive group action $G$, where any two points $p, q \in\M$ can be connected by some group element, i.e. there exists $g\in G$ such that $p = g\cdot q$. In this setting, the manifold is called a homogeneous space which essentially equivalent to a quotient group $G/G_p$ where $G_p$ is the stabilizer of the group action at $p$. However, the general manifolds considered in this paper often do not have an associated transitive group action. Therefore, it is still necessary to consider a new method to define convolution on manifolds without group action structure. 

\begin{table*}[htp]
\begin{center}
\vspace{-0.5cm}
\begin{center}
\resizebox{14cm}{1.4cm}{
    \begin{tabular}{| l || c | c | c | c  | c | c |}
    \hline
                            Method   & Filter Type & Support  & Extraction & Directional   & Transferable & Deformable \\ 		\hline
    Spectral~\cite{bruna2013spectral} & Spectral    & Global   & Eigen          & Yes            & No         & No \\
    TFG~\cite{dong2015sparse} & Spectral    & Global   & Eigen          & Yes            & No         & No \\
    WFT~\cite{shuman2016vertex} & Spectral    & Local   & Windowed Eigen          & Yes            & No         & No \\
    GCNN~\cite{masci2015geodesic}     & Patch       & Local    & Variable       & No             & Yes         & Yes \\
    ACNN~\cite{boscaini2016learning}  & Patch       & Local    & Fixed          & Yes           & Yes         & No \\
    \hline
    \textbf{PTC} & Geodesic    & Local    & Embedded & Yes            & Yes         & Yes \\
    \hline
    \end{tabular}
    }
\end{center}
\label{CompTable}
\caption{Comparison on different generalizations of convolutional operator on general manifolds.}
\end{center}
\end{table*}

In the Euclidean setting, convolution operators that are frequently used in practice have compactly supported filters. Furthermore, they are directionally aware, deformable and can be easily transferred from one signal domain to another. Previous attempts to generalize the convolution operator on manifolds have failed to preserves one or more of these key properties. In this paper, we propose a new way of defining the convolution operation on manifolds based on parallel transportation. We shall refer to the proposed convolution as the parallel transportation convolution (PTC). The proposed PTC is able to preserve all of the aforementioned key proprieties of Euclidean convolutions. This spatially defined convolution operation enjoys flexibility of conducting isotropic or anisotropic diffusion, and it also enables us to perform wavelet-like operations as well as defining convolutional neural networks on manifolds. In addition, the PTC can be reduced to the Euclidean convolutions when the domain is flat. Therefore, the PTC is a natural generalization of Euclidean convolution operators on manifolds.

To be more precise, we seek a convolution operator of the form:
\begin{equation}\label{ManConv}
(f *_{\M} k) (x) := \int_{\M} k(x,y)f(y) d_{\M}y.
\end{equation}
In the Euclidean case, $x-y$ essentially represents the direction from $x$ to $y$, while on manifold such a vector can be understood as a tangent direction at $x$ pointing to $y$. The crucial idea of PTC is to define a kernel function $k(x,y)$ which is able to encode $x-y$ using a parallel transportation that naturally incorporates the manifold structure.

Table \ref{CompTable} compares the proposed PTC with previous efforts. Since the group action methods are limited to homogenous spaces, which do not fit our objective of designing convolution on general manifolds, we do not include these methods in the table. A method is called directional if the filters are able to characterize non-isotropic features of the data. A method is transferable if the filters can be applied to manifolds with different LB eigensystems. Finally, a technique is said to be deformable if large deformations in the manifold (i.e. those which change properties such as curvature or local distances) do not drastically affect the convolution.


The rest of this paper is organized as follows. In section \ref{sec:Math}, we discuss necessary mathematical background of differential manifolds and parallel transportation of vector fields on manifolds. Then, we introduce the proposed PTC on manifolds in section \ref{sec:PTC}. We also discuss implementation details and how to design convolutional neural networks on manifolds using the proposed PTC. In section \ref{sec:experiments}, several numerical experiments illustrate the effectiveness of the proposed method. Finally, concluding remarks are made in section \ref{sec:conclusion}.

\section{Mathematical Background}
\label{sec:Math}
In this section, we discuss some background of differential manifolds and parallel transportation. These provide theoretical preparation of the proposed convolution on manifolds.

\subsection{Manifolds, Tangent Spaces and the Exponential Map}

Let $\M$ be a two dimensional differential manifold associated with a metric $g_{\M}$. For simplicity we assume that $(\M,\delta_M)$ is embedded in $\RR^3$. We write the set of all tangent vectors at a point $x\in\M$ as $\T_x\M$ referred as the tangent plane of $\M$ at $x$. The disjoint union of all tangent planes, $\bigcup_{x} \{(x,v)\in\M\times \RR^3~|~x\in\M, v\in\T_x\M\}$, forms a four dimensional differential manifold called the tangent bundle $\T\M$ of $\M$. A vector field $X$ is a smooth assignment $X: \M\rightarrow\T\M$ such that $X(x)\in\T_x\M, ~\forall x\in\M$. We denote all smooth vector fields on $\M$ as $C^{\infty}(\M, \T \M)$.

Let $\T_{x,\delta} \M = \{v\in \T_{x}\M ~|~ \langle v,v\rangle_{g_{\M}} \leq \delta\}$ be a $\delta$-neighborhood of the tangent space at a given point $x$. The {\it exponential map}, $\exp: \T_{x,\delta} \M \rightarrow \M_{x,\delta}$, maps vectors from the tangent space back onto a nearby region $\M_{x,\delta}$ of $x$ on the manifold. Formally, given $v\in \T_{x,\delta}\M$ there exists a unique geodesic curve $\gamma$ with $\gamma(0)= x$ and $\gamma'(0) = v$ such that $\exp_{x}(v) = \gamma(1)$.
Note that this map is defined in the local neighborhood where the differential equation: $\gamma'(0) = v$ with initial condition $\gamma(0) = x$ has a unique solution. The size of this neighborhood depends on the local geometry of the manifold. In fact,  the exponential map defines a one-to-one correspondence between $\T_{x,\delta}\M$ and $\M_{x,\delta}$ if $\delta$ is smaller than the injective radius of $\M$~\cite{kobayashi1969foundations,chavel2006riemannian}. Since this map is a bijection, there is a natural inverse which we denote as $\exp_{x}^{-1}: \T_{x,\delta} \M \rightarrow \M$.

\subsection{Parallel Transportation}
\label{subsec:PT}
In differential geometry, parallel transportation is a way of translating a vector, based an affine connection, along a smooth curve so the resulting vector is `parallel'. An affine connection translates the tangent spaces of points on a manifold in a way that allows us to differentiate vector fields along curves. Formally, an {\it affine connection} is a bilinear map $\nabla: C^{\infty}(\M, \T \M) \times C^{\infty}(\M, \T \M) \rightarrow C^{\infty}(\M)$, such that for all smooth functions $f, g$ and all vector fields $X,Y, Z$ on $\M$ satisfy:
\begin{equation}
\left\{\begin{array}{c}
\nabla_{fX + gY} X = f \nabla_X Z + g \nabla_Y Z \qquad \quad \\
\nabla_{X} (aY + bZ) = a \nabla_X Y + b \nabla_X Z  \quad a, b \in \RR \\
\nabla_X (fY) = df(X)Y+f\nabla_X Y  \qquad
\end{array}
\right.
\end{equation}
In particular, an affine connection is called the Levi-Civita connection if it is torsion free ($\nabla_X Y - \nabla_Y X = [X,Y]$) and compatible with the metric ( $X\langle Y,Z\rangle_{g_\M} = \langle\nabla_X Y,Z\rangle_{g_\M} + \langle Y,\nabla_X Z\rangle_{g_\M}$).
In this case, the transport induced from the connection preserves the length of the transported vector and the angle it makes with each side.

A curve $\gamma:[0,\ell] \rightarrow \M$ on $\M$ is called geodesic if $\nabla_{\dot \gamma(t)} \gamma(t) = 0$. More precisely, using local coordinate system, we can write $\displaystyle \dot \gamma (t) = \sum_{i=1}^2 \frac{dx^i}{dt}\partial x^i$, then plugging in the covariant derivative leads to the following ordinary differential equation for a geodesic curve:
\begin{equation}\label{Geodesic Equation}
\frac{d^2x^k(t)}{dt^2} + \sum_{i,j=1}^2 \Gamma^k_{ij}(t) \frac{dx^i(t)}{dt} \frac{dx^j(t)}{dt} = 0, \qquad k = 1, 2
\end{equation}
where $\Gamma^k_{i,j}$ is the Christoffel symbols associated with the local coordinate system.
For any two points $x_0$ and $x_1$ on a complete manifold $\M$, there will be a geodesic $\gamma:[0,\ell] \rightarrow\M$ connecting $x_0$ and $x_1$.
A vector field $X(t)$ on $\gamma(t)$ is called {\it parallel}  if $\nabla_{\dot \gamma} X = 0$. Therefore, given any vector $v \in \T_{x_0} \M$,  we can transport $v$ to a vector $v'$ in $\T_{x_1}\M$ by defining $v' = X(\ell)$ from the solution of the initial value problem $\nabla_{\dot \gamma(t)} X(t) = 0$ with $X(0) = v$.
In other words,
If we write $X(t) = \sum_{i=1}^2 a^i(t)\partial x^i$, the problem of solving $X$ reduces to find the appropriate coefficients $\{a^k(t)\}$ satisfying the parallel transport equation. This can be written as the following first order linear system:
\begin{equation}
\left\{\begin{array}{c}
\displaystyle \dfrac{d a^k(t)}{dt} +  \sum_{i,j=1}^2 \dfrac{d \gamma^i}{dt} a^j(t)\Gamma ^k_{ij} = 0, \quad k = 1, 2 \vspace{0.2cm}\\
\sum_{i=1}^2 a^i(0) \partial x^i = v
\end{array}\right.
\end{equation}
Solving this equation finds a parallel vector field $X$ along $\gamma(t)$ which provides parallel transportation of $v = X(0) \in\T_{x_0}\M$ to $X(\ell)\in\T_{x_1}\M$. We denote the parallel transportation of a vector from $x_0$ to $x_1$ along the geodesic as $\PT_{x_0}^{x_1}:\T_{x_0,\delta}\M \rightarrow\T_{x_1,\delta}\M$.

\section{Parallel Transport Convolution (PTC)}
\label{sec:PTC}
In this section, we introduce parallel transport convolution on manifolds which provide a fundamental important building block of designing convolutional neural networks on manifolds. After that, we discuss numerical discretization of PTC and PCTNet on manifolds.
\subsection{Mathematic definition of PTC}
Unlike one-dimensional signals or images whose base space is shift invariant, geometric objects modeled as curved manifolds do not have shift-invariance. This is an essential barrier to adopt CNN to conduct learning on manifolds and graphs except for a few recent work where convolution is defined in the frequency space of the LB operator~\cite{bruna2013spectral,shotton2013real,rodola2014dense,rustamov2013wavelets}. These methods only manipulates the LB eigenvalues by splitting the high dimension information to LB eigenfunctions. Limitations include that it is always isotropic due to the LB operator and can only approximate the even order differential operators~\cite{dong2015sparse}. In addition, there is another recent method discussed in~\cite{masci2015shapenet}, in which convolution is directly considered on the spatial domain using local integral on geodesic disc although it does not involve manifold structure as transportation on manifold is not considered. The lack of an appropriate method of defining convolution on manifolds motivates us to introduce the following way of defining convolution on manifolds through parallel transportation. This geometric way of defining convolution naturally integrates manifold structures and enables futher learning tasks.

Let $\M(x_0, \delta) = \{y \in \M \ | \ d_{\M}(x_0,y) \leq \delta \}$ and $k(x_0,\cdot): \M(x_0,\delta)\rightarrow \RR$ be a compactly supported kernel function center at $x_0$ with raduis $\delta$. We assume $k(x_0,y) = 0$ for $y \notin \M(x_0,\delta)$ and require the radius of the compact support parameter $\delta$ be smaller than the injective radius of $\M$ to guarantee the bijectivity of the exponential map. It is important to remark that $k(x_0,\cdot)$ can be determined by users, be designed for specific applications, or be optimized in the later exploration of parallel transport convolutional neural networks.
Our idea of defining convolution on manifolds is conducted by transporting the compactly supported kernel $k(x_0,\cdots)$ to every other point on $\M$ in the way of reflecting the manifold geometry.
More specifically, given any point $x\in\M$, we first construct a vector field transportation $\PT_{x_0}^x:\T_{x_0,\delta}\M \rightarrow \T_{x,\delta}\M$ using the parallel transportation discussed in Section \ref{subsec:PT}. Then $k(x_0,\cdot)$ can be transported on $\M$ as:
\begin{equation}
\label{eqn:kernel}
\begin{aligned}
k(x,\cdot):\M_{x,\delta} &\rightarrow  \RR \\
 y & \mapsto  k\left(x_0, \exp_{x_0}\circ (\PT_{x_0}^{x})^{-1}\circ \exp_x^{-1}(y)\right)
\end{aligned}
\end{equation}
Note that the above definition is an analogy of the convolution kernel in the Euclidean space. Here, the exponential map $\exp_x^{-1}(y)$ mimics the vector $x - y$, and $\PT_{x_0}^x$ is a generalization of the shift operation in the Euclidean case where $x_0-y$ is shifted to $x-y$. In fact, it can be easily checked that the above definition is compatible with Euclidean case by setting the manifold $\M$ to be $\RR^2$.

By plugging \eqref{eqn:kernel} into \eqref{ManConv}, we can now formally define the {\it parallel transport convolution} operation of $f$ which a filter $k$, centered at $x_0$:
\begin{equation}
\begin{split}\label{ManConvRot}
(f *_{\M} k) (x) := \int_{\M} f(y)~k(x,y) d_\M y =  \\
\int_{\M} f(y) ~k\left(x_0, \exp_{x_0}\circ (\PT_{x_0}^{x})^{-1}\circ \exp_x^{-1}(y)\right) \ d_{\M}y
\end{split}
\end{equation}
As natural extensions, this approach can also be used to define dilations, reflections and rotations by simply manipulating the reference vector $\exp^{-1}_{x}(y)$. More specifically, shrinking or expanding the kernel by a factor of $s$ is defined by multiplying the lengths of the vectors in the tangent space by $s$. If $s$ is chosen to be negative then the kernel is reflected and dilated by a factor of $|s|$. Similarly, rotating the kernel is defined as conducting a rotation operator $R_{\theta}$ to the reference vectors on the tangent plane. In summary, the scaling of $k$ by $s$ with a rotation of $\theta$ is defined as:
\begin{equation}\label{RotandDil}
k_{s,\theta}(x,y) := \frac{1}{C_{x}} k\Big(x_0,\ \exp_{x_0}\circ (\PT_{x_0}^{x})^{-1}(s\  R_{\theta} \ exp^{-1}_{x}(y)\big)\Big)
\end{equation}
where $R_\theta$ is a rotation matrix and $\displaystyle \frac{1}{C_{x}}$ is a normalization constant that can be used to preserve volume of the kernel.

\subsection{Numerical Discretization}
\begin{figure}
 \centering
\includegraphics[width=.4\textwidth]{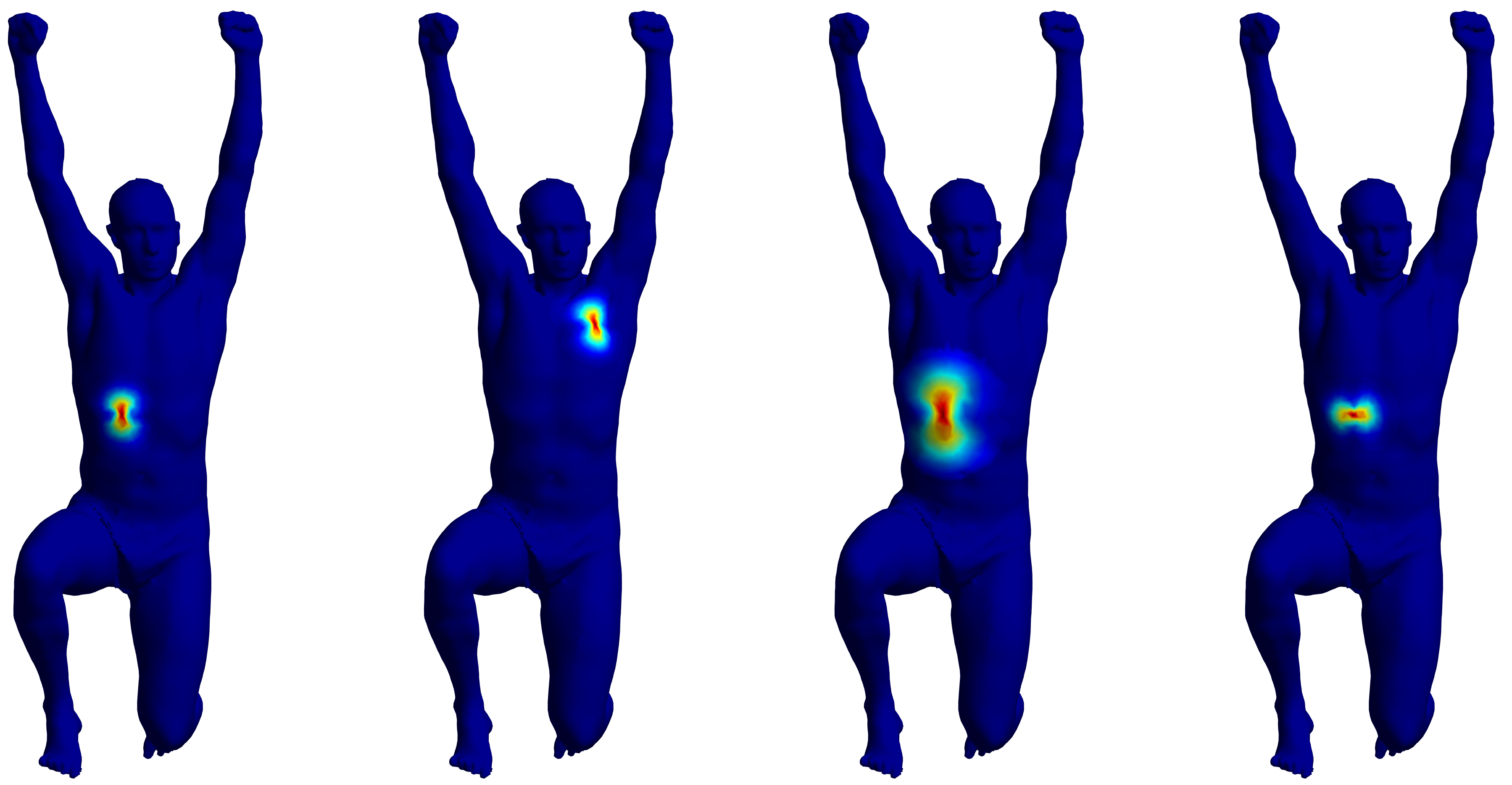}\\
{\centering (a)\hspace{1.7cm} (b) \hspace{1.7cm} (c) \hspace{1.7cm} (d)}
\caption{A compactly supported kernel (a) is transported on a manifold from the FAUST data set~\cite{bogo2014faust} through translation (b), translation + dilation (c) and translation + rotation (d).}
  \label{fig:ConvKernel}
\end{figure} 
We represent a two-dimensional manifold $\M$ using triangle mesh $\{V, E, T\}$. Here $V = \{v_i \in \RR^3\}_{i=1}^n$ denotes vertices and $T = \{\tau_s \}_{s=1}^l$ denotes faces. First we compute the geodesic distance function from $x_0$ to every other point by solving the Eikonal equation $|\nabla_\M D(x) | = 1$ using the fast marching method \cite{sethian1996fast,kimmel1998computing}. Next we calculate $\nabla_\M D$ and its orthonormal direction on each triangle $\tau_s$. Together with the face normal direction $\vn_{s}$,
for each triangle $\tau_s$, we construct a local orthonormal frame $\mathfrak{F}_s = \{\vb_{s}^1,\vb_{s}^2,\vn_{s}\}$ where $\vb_{s}^1,\vb_{s}^2$, reflecting the intrinsic information, are tangent to $\tau_s$,  and $\vn_{s}$, reflecting the extrinsic information, is orthogonal to $\tau_s$. For an edge adjacent with $\tau_s$ and $\tau_t$, we write $R_{st}$ as an orthonormal transition matrix such that $R_{st}\mathfrak{F}_t = \mathfrak{F}_s$. Then any vector in $\mathrm{Span}\{\vb_{s}^1,\vb_{s}^2\}$ can be transported to $\mathrm{Span}\{\vb_{t}^1,\vb_{t}^2\}$ using the transition matrix $R_{st}$. This can be viewed as a discretization of connection and used to transport a vector on the tangent space of one given point to all other points.
The compatibility condition of all $R_{st}$ discussed in~\cite{wang2012linear} can guarantee  that no ambiguity will be introduced in this way.


After the transportation is conducted, the convolution kernel can be transported to a new point by interpolating the transported vectors in the local tangent space at the target point.
Computationally, we define a sparse matrix $K$ where the $i^{th}$ column is the transportation of the kernel to the $i^{th}$ vertex. Thus, we have the following definition of discrete parallel transport convolution:
$(f *_{\M} k) (x) :=  K^T \textbf{M} F$
where $F$ is column vector representation the function $f$ at each vertex and $\textbf{M}$ is the mass matrix. Note that once we have found the vector field of the geodesic equation, the transportation of the kernel to each new center and multiplication with $F$ is independent and can therefore be parallelized efficiently. 
Figure~\ref{fig:ConvKernel} illustrates the effect of the proposed method of transporting a kernel function on a manifold. This result shows that the proposed method produce an analogy of the behavior of a kernel function $k(x-y)$ operating in the Euclidean domain. More importantly, we would like to emphasize that number of freedoms in our PTC is essentially the same as the classical convolution on Euclidean domain. This makes our method has much less number of parameters as those used in the patch based methods~\cite{masci2015geodesic}. In addition, PTC can be computed very efficiently using sparse matrices product once the interpolation matrices and mass matrices have been precomputed. We provide detailed implementation about sparse matrices multiplication of PTC in the appendix. 

\subsection{Convolutional neural networks on manifolds through PTC}
Using the proposed PTC, we can define convolutional neural networks on manifolds. We shall refer these network as PTCNets.
Similar as CNNs on Euclidean domains, a PTCNet consists of an input and an output layer, as well as multiple hidden layers including fully connected layers, nonlinear layers, pooling layers and PTC layers listed as follows.
\begin{itemize}[leftmargin=0.4cm]
\item Fully connected layer: $f_i^{out}(x) = \sum_{j=1}^{N} w_{ij} f^{in}_j(x),  \quad i = 1,\cdots,L$. This layer connects every neuron in one layer to every neuron in the previous layer.
The coefficient matrix $(w_{ij})$ parametrizes this layer and will be trained by a training data set.
\item ReLu layer: $f^{out}_i(x) = \max\{0,f_i^{in}(x)\}, \quad i = 1,\cdots,L$. This is a fixed layer applying the nonlinear Rectified Linear Units function $\max\{0,x\}$ to each input.
\item PTC layer: $f^{out}_{i,\alpha}(x) = \int k_\alpha(x,y) f^{in}_i(y) ~\mathrm{d} y \approx K_{\alpha}\textbf{M} F^{in}_i, \quad \alpha = 1,\cdots, m$.
This layer applies the proposed PTC to the input, passes the result to the next layer.
Each $k_\alpha$ is determined by the proposed PTC on manifolds with an initial convolution kernel $k_\alpha(x_0,\cdot)$, which parametrize the parallel transport convolution process and will be trained based on a training data set. For certain applications with a moderate size of training set, more structured initial kernel might be needed. In this case, we can control $k_\alpha$ by a sequence of rotation in the tangent space, which can reduce the number of free parameters and save computation time. Detail on memory efficient implementation of this layer can be found in the appendix
\end{itemize}
Therefore, it is straightforward to adapt established network architectures in Euclidean domain cases to manifolds case as the only change is to replace traditional convolution by PTC. In addition, back-propagation can be achieved by taking derivation of $K$. The compact support of the convolution kernel is represented as a sparse matrix which makes computation efficient.


Thus far we have only considered transportation along the geodesic. In practice we can compute the parallel transportation along any given vector field. For some applications it may be more natural to use another vector field. To do so we follow the same process except using this new vector field to form the first basis vector in $V$. This can be extremely beneficial in dealing with areas in which our geodesic vector field has a singularity. Around the singularity the direction of the vector field is often highly variable. We can simply define another vector field which is more regular in this area (but may have singularities else where) to analyze information near the singularity in the first field. The problem of designing and controlling the singularities of vector fields on surfaces is a well studied problem for which many approaches already exist (see \cite{DeGoe2015vector} for a review of such techniques). It is important to note that if we would like our the results of our training to be generalizeable (i.e. when working with multiple domains) then we need to the vector fields to be generalizeable as well. For this reason using geodesic distances from canonically chosen points is a natural choice.

\section{Numerical Experiments}
\label{sec:experiments}
To illustrate the effectiveness of the proposed PTC,  we conduct numerical experiments including processing images on manifolds using PTC, classifying images on manifolds using PTCNets and learning features on manifolds for registation. All numerical experiments on MNIST data were implemented in MATLAB on a PC with a 32GB RAM and 3.5GHz CPU, while the final experiment was implemented in Tensorflow with a NVIDA GTX 1080 T graphics card.
We remark that these experiments aim to demonstrate capabilities the proposed PTC for manipulating functions on curved domains by naturally extending existing wavelet and learning methods from Euclidean domains to curves domains. It is by no means to show that the experiments achieve state-of-the-art results on euclidean problems. In our future work, we will show its further applications of comparing, classifying and understanding manifold-structured data by combing with recent advances in deep learning architectures.
\subsection{Wavelet-like Operations}
\begin{figure}[htp]
\begin{center}
\includegraphics[width=1\linewidth]{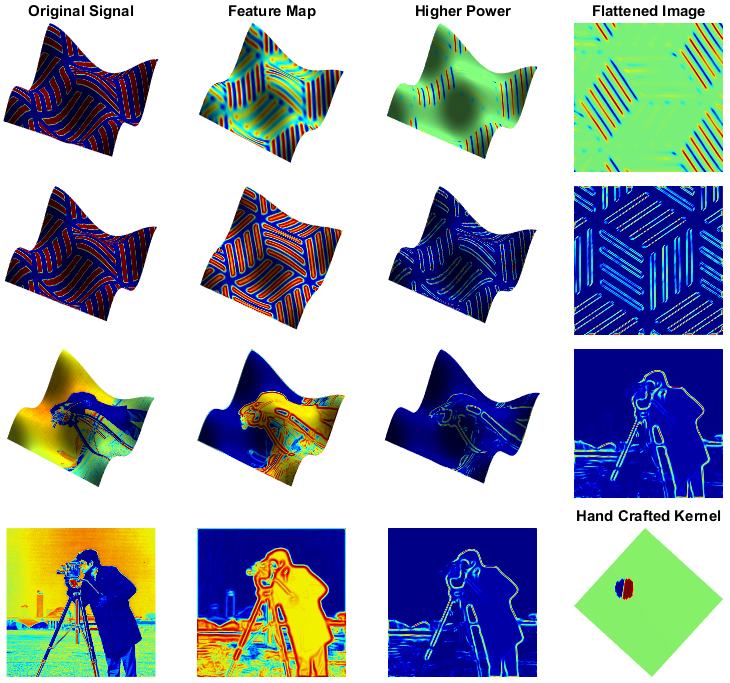}
\end{center}
\caption{First Row: Convolutions with out rotation on test image. Second Row: Convolutions with rotation on test image. Third Row: Convolutions with rotation on a cameraman image. Fourth row: Traditional Euclidean convolution and the edge detector used in PTC.}
\label{fig:wavelet}
\end{figure}
In the first experiment, we demonstrate the effectiveness of our approach by performing simple signal processing tasks on manifolds. Then we  compare the PTC results to those produced by traditional techniques applied to Euclidean domains. First we apply PTC with a hand crafted edge detection filter to images on a manifold. By convolving this filter with the input image, we obtain an output feature function whose higher values indicate similarity to the predefined edge. In the first row of Figure~\ref{fig:wavelet}, it is clear that the proposed convolution successfully highlights the edges with similar orientation of the input filter. In the second row of Figure~\ref{fig:wavelet}, we allow additional rotations as we discussed in \eqref{RotandDil}. We observe that the additional rotation flexibility can reliably capture all of the edges regardless of orientations. This illustrates the directional awareness of our method. Furthermore, we apply this edge detector using PTC to a more realistic problem in the third row of Figure~\ref{fig:wavelet}. It shows that the results are very close to those produced in an analogous Euclidean setting (fourth row). In the third column, we show the feature map raised to the fifth power for better contrast and the last column shows a flattened version for easier visualization.

\subsection{Single Manifold MNIST}
\label{subsec:SingleMfd}
\begin{figure}[htp]
\begin{center}
\resizebox{0.47\textwidth}{0.4in}{
\vspace{-0.6cm}
  \begin{tabular}{ c | l |c|c | }
 \cline{2-4}
\multirow{ 5}{*}{\includegraphics[width=.12\textwidth]{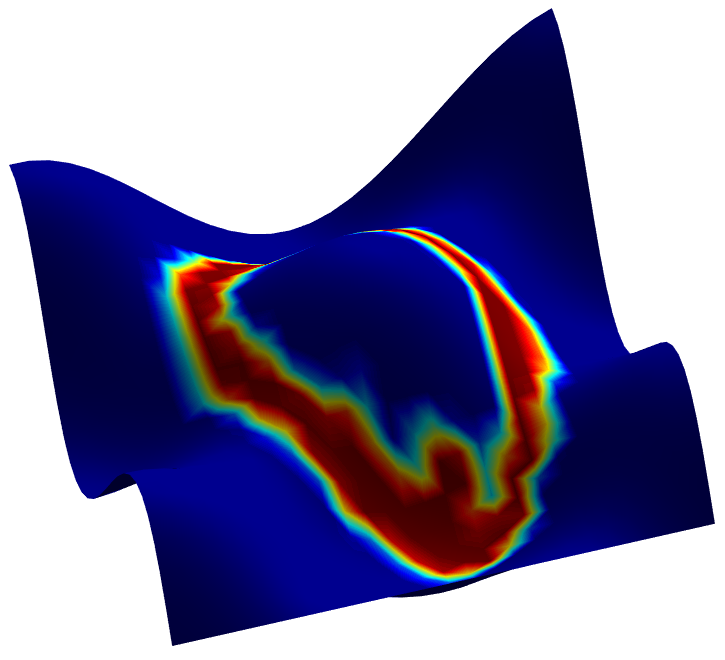}} &  Network  & Domain & Success Rate \\ \cline{2-4}
   & Traditional              & Euclidean & \textbf{98.85} \\ \cline{2-4}
    & \textbf{Flat PTCNet} & Euclidean  & 98.10  \\ \cline{2-4}
   & Spectral             & Manifold  & 95.35 \\ \cline{2-4}
   &  \textbf{PTCNet}   & Manifold     & 97.96  \\ \cline{2-4}
  \end{tabular}}
\end{center}
\caption{Comparison of our PCTNet to Euclidean case and a spectral based method on a single manifold.}
\label{tab:SingleMNIST}
\end{figure}

In this test, we conduct experiments to demonstrate the effectiveness of PTCNets to handle signals on manifolds. The most highly celebrated early applications of CNNs was the recognition of hand written digits \cite{lecun1998gradient}. We map all MNIST data to a curved manifold plotted in the left image of Figure~\ref{tab:SingleMNIST}. We use a simple network architecture consisting of a single convolution layer with 16 filters followed by a ReLu non-linear layer and then a fully connected layer which outputs a 10 dimensional vector of predictions. We apply this network architecture to four scenarios including MNIST data on a Euclidean domain using traditional convolution, MNIST data on a Euclidean domain using PTC,  MINST data on a curved domain using PTC, and MINST data on the same curved domain using spectral convolution.

Each network is implemented in MATLAB using only elementary functions and  is trained using batch stochastic gradient descent with batch size 50 and a fixed learning rate $\alpha = 10^{-3}$. We also use the same random seed for the batch selection and the same initialization. We choose such a simple training regime in order to make the effects of different convolution operations as clear as possible. We measure the results by the overall network error after 5,000 iterations.

The table~ in Figure~\ref{tab:SingleMNIST} shows the accuracy of the traditional CNN on a flat domain, a spectral net applied to a simple manifold as well as our network applied to both a Euclidean domain (Flat PTCNet) and the manifold. Similar performance of Flat PTCnet to traditional CNN illustrate that our method is an appropriate generalization of convolution from flat domains to curved domains. In addition, we observe that our method out preforms the spectral network for this classification task on a curved domain.

\subsection{Multi-Manifold MNIST}
\begin{figure}[htp]
\begin{center}
  \begin{tabular}{ |c | c | }
  \hline
 Training      &  Success Rate  \\ \hline
 Spectral      & 88.50        \\ \hline
 Single Manifold     & 95.65  \\ \hline
 \textbf{Multiple Manifolds} & \textbf{97.32} \\ \hline
  \end{tabular}
\end{center}
\centering
\includegraphics[width=.95\linewidth]{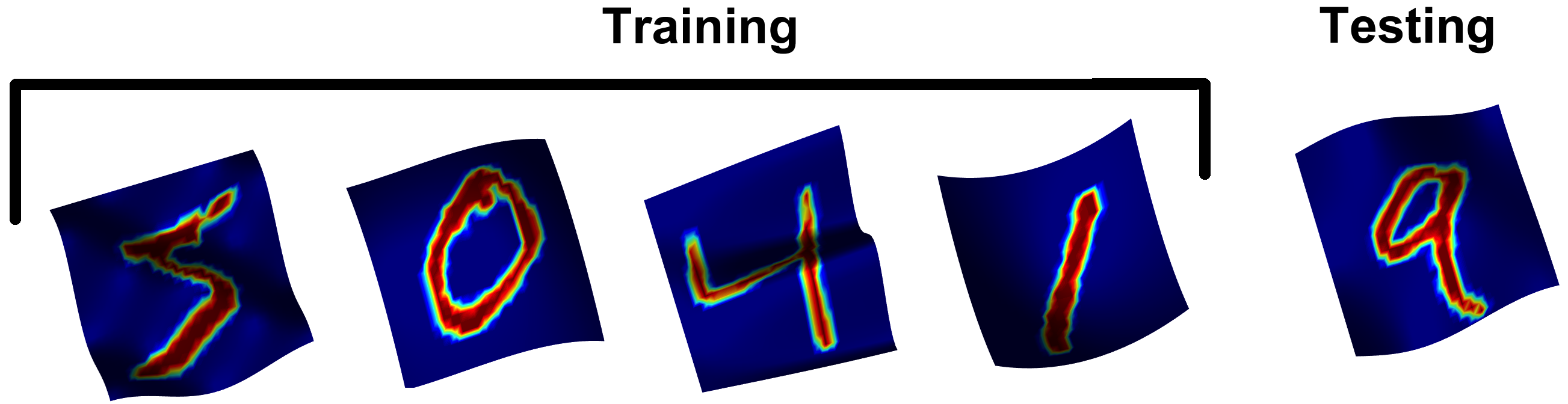}
\caption{Top table: Comparison of results from learning on single and multiple domains and then testing on a new manifold. Bottom: Manifolds used for muli-mainfold tests. The first four are used for training and the last is used for testing.  }
\label{fig:MultiMNIST}
\end{figure}

One of the advantages of our method is that filters which are learned on one manifold can be applied to different domains. The spectral convolution based methods do not have this transferability as different domains are unlikely to share the same eigensystem. In this experiment, we first directly apply the network learned by the PTCNet and Spectral networks from Section~\ref{subsec:SingleMfd} to a new manifold. As we illustrate in the first two rows of the table in Fig.~\ref{fig:MultiMNIST}, the accuracy of the spectral convolution based method is dramatically reduced since the two manifolds have quite different eigensystems. However, our PTCNet can still provide reasonable accurate results since the underlying geodesic vector fields of these manifolds is more stable to deformations than eigensystems are.

Furthermore, we conduct a new experiment in which we train our PTCNet on a variety of manifolds and test on a different manifolds as showed in the bottom picture of Figure \ref{fig:MultiMNIST}, where the first four manifolds are used as training domains, and the fifth one is used for testing. From these pictures, it is clear that the training manifolds are quite different and therefore the spectral methods and definitions of convolution which require curvature to set their direction \cite{boscaini2016learning} cannot be applied to these problems. However the geodesic vector fields  of the manifolds are quite similar and therefore filters learned through our technique should apply to the new problem. As we can see in the last row of the Table in Figure~\ref{fig:MultiMNIST}, the network achieves a 97.32$\%$ success rate since training on multiple manifolds allows PTC network to learn greater invariance to local deformation in the metric, which enables great transferability.

\subsection{Singularities of vector fields}
In each of the previous experiments the vector field used to translate the convolutional kernels is choosen to be the gradient of the geodesic from one corner of the manifold. Although our convolution is well defined everywhere on these manifolds, the filters may be more variable near this singularity. To investigate the effects that these singularities may have on,  we next test our network using  different types of vector fields. PTC1 uses the vector field chosen as in the previous experiments. PTC2 uses a vector field with a singularity in the center of the domain. The next test (PTC3) has two separate vector fields each with a singularity at different point on the interior of the domain. For this test, half the kernels are assigned to one vector field and half to the other. The last test uses four vector fields, each with a singularity at a different point on the interior of the manifold. Table~\ref{table:Sing} shows the results of using these vector fields on the single and multiple manifold problems described previously. We observe that the presence of singularity can negatively effect the performance, while using multiple vector fields can overcome these difficulties.
\begin{table}[htp]
\begin{center}
\resizebox{0.48\textwidth}{0.45in}{
  \begin{tabular}{ | c || c | c | c | c | }
  \hline
    Implementation      &  VF & Sings per VF   & Single:  SR  & Multi:  SR   \\ \hline
    Spectral                 & - & -     & 92.10  & 88.50   \\ \hline
    PTC1                     & 1 & 0     & \textbf{96.36}  & \textbf{97.32}   \\ \hline
    PTC2                     & 1 & 1     & 94.92  & 94.51   \\ \hline
    PTC3                     & 2 & 1     & 95.89  & 95.02  \\ \hline
    PTC4                     & 4 & 1     & 96.01  & 95.28   \\ \hline
  \end{tabular}
  }
\end{center}
\caption{Success rate (SR) comparison of several of our networks on a single (the $4^{th}$ coloum) and on multiple manifolds (the $5^{th}$ coloumn).}
\label{table:Sing}
\end{table}

\subsection{Feature Learning for Shape Registration}
\begin{figure*}[t]
\begin{minipage}{0.55\linewidth}
\centering
\includegraphics[width=1\linewidth]{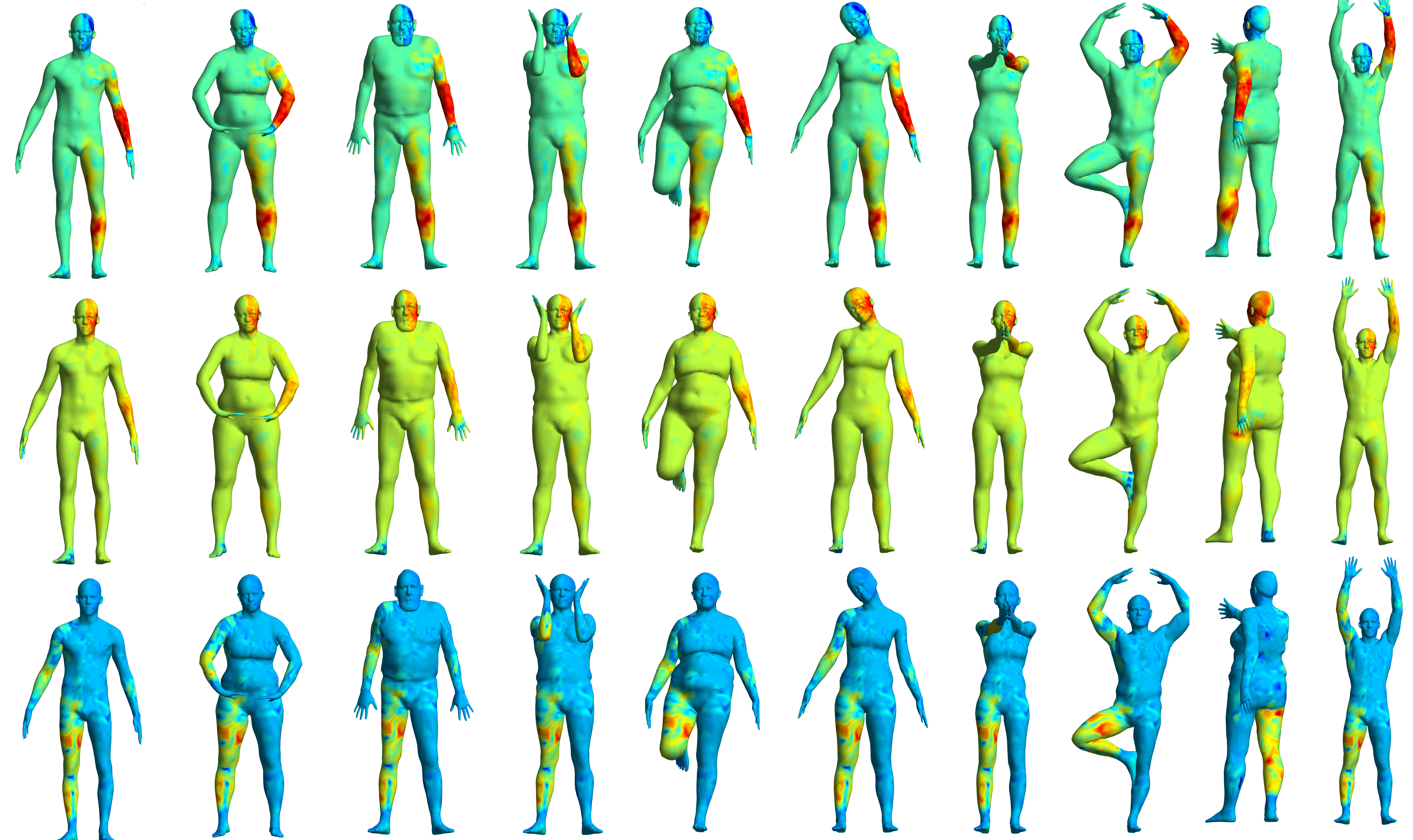}
\end{minipage}\hfill
\begin{minipage}{0.45\linewidth}
\includegraphics[width=1\linewidth]{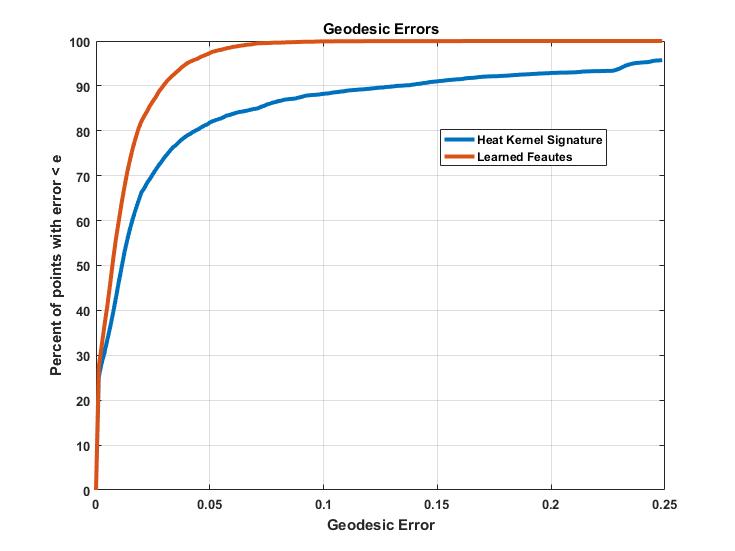}
\end{minipage}
\caption{Left: Example feature functions for shape correspondence on the Faust dataset. Right: geodesic errors in predicted correspondence.}
\label{fig:FaustFeats}
\end{figure*}
One important application of convolution neural networks in shape processing is the creation of geometric features \cite{bronstein2017geometric}. The goal of these networks is to output descriptor functions, $F:(\M) \rightarrow  \RR$, which accurate describe the local and global geometry of a manifold. In this section we implement a network based on the 'ShapeNet2' architecture original presented in \cite{masci2015geodesic} for shape registration, substituting in our proposed definition of convolution. In this network we input a 150 dimensional geometry vector into a vector connected layer which linearly combines these input features into a 16 dimensional signal. This signal is then passed through two layers of PTC  (each followed by a Relu non-linearity) with 16 filters in each layer. The final features are the output of the second convolution layer. The network is trained by minimizing the following triplet loss:
\begin{equation}
\begin{split}
L(S;\Theta) &= 
\sum_{x_1,x_2 \in S \times S} ||F(x_1;\Theta) - F(x_2;\Theta)||^2 
\\
&+ \lambda \sum_{P \in \Pi}(\mu_1 - || F(x_1;\Theta) -  F(x_3;\Theta))||)^2 
\end{split}
\end{equation}
where $\{x_1,x_2\}$ are similar pairs of shapes, $\{x_1,x_3\}$ are dissimilar and $\mu$ is the user parameter representing the margin. We evaluated this model the Faust dataset which contain 100 real world scans (each with $n = 6890$ points) of 10 individuals in 10 poses~\cite{bogo2014faust}. We use the first 80 figures for training, 10 for validation, and 10 for testing. Using the sparse matrix operations described in the appendix, each forward and backward propagation through a two layer network, defined on a mesh containing 6890 points, can be calculated in less than half a second. The whole training process is completed in 8 hours using the ADAM algorithm~\cite{kingma2014adam}.  Figure \ref{fig:FaustFeats} shows three of the output feature functions across different individuals in the dataset, where the first 7 individuals (10 surfaces for each individual) are used for training, the $8^{th}$ and $9^{th}$ individuals are used for validation, and the last individual is used for testing. These consistent features lead to satisfactory registration results by simply conducting the nearest point search in the feature space.  Figure \ref{fig:FaustFeats} shows our registration performance, measured by the geodesic error between the predicted correspondence and the actual correspondence, compared to error from use the heat kernel signatures which were used as input layer.

%

\section{Conclusion}
\label{sec:conclusion}
In this paper, we propose a generalization of the convolution operation on smooth manifolds using parallel transportation and discuss its numerical implementation. Using the proposed PTC, we have preformed wavelet-like operation of signals and built convolutional neural networks on curved domains. Our numerical experiments have shown that the PTC can preform as well as Euclidean methods on curved manifolds, and is capable of including directional awareness, handling problems involving deformable manifolds, in particular, learning features for deformable manifolds registration. In our future works, we will apply our PTC to different applications of comparing, classifying and understanding manifold-structured data by combining with recent advances of deep learning architectures.

\section*{Appendix: Efficient computation of PTC layers}
Since the limitation of spare matrix product implementation in TensorFlow and PyTorch, we use the following method to implement the proposed convolution. More specifically, we consider a mesh with $n$ points, a signal with $q$ channels $F = (F_1,\cdots,F_q)\in\RR^{n\times q}$ and $p$ filters  each of which has $q$ input channels denoted $\textbf{K} = \{K_{11},\cdots,K_{1p},\cdots,K_{q1},\cdots,K_{qp}\}$. We would like to compute convolution $F\star \textbf{K} =  \sum_{i=1}^{q} F_i \star K_{ij} \in\RR^{n\times p}$. Given a mesh with the mass matrix $M$, we write $I_i$ as the index set of the neighborhood of the $i$ point and denote $W_i\in \RR^{|I_i| \times k}$ the parallel transportation operation to the $i$-th point. 
The following method provides a fast, memory efficient implementation of PTC convolution in TensorFlow and PyTorch. 

We write $Z_i = F_i ^T M \in \RR^{n\times 1} , i = 1,\cdots,q $  and let $L = \sum_{i}|I_i|$. We define $\textbf{Z}_i$ as a $L\times L$ sparse matrix whose support at the $k$-th row is provided by $I_k$ with value $Z_i(I_k)$, formally we write:
$$\textbf{Z}_i = \begin{pmatrix}
Z_i(I_1) \\
Z_i(I_2) \\ 
\vdots   \\  
Z_i(I_n) \\
\end{pmatrix},~
\textbf{Z}= 
\begin{pmatrix}
\textbf{Z}_1   & &   &  &   \\ 
& \textbf{Z}_2 & & \bigzero &  \\ 
& &\ddots  & &  \\ 
& \bigzero & & \ddots    &    \\ 
&  &  & &   \textbf{Z}_q
\end{pmatrix}
$$
In addition, we define
$$
\textbf{W} = 
\begin{pmatrix}
W_1 \\ 
W_2 \\ 
\vdots \\ 
\vdots \\
W_n 
\end{pmatrix}, 
\bar{\textbf{W}} = 
\begin{pmatrix}
\textbf{W}K_{11}   & \cdots &  \textbf{W}K_{1p} \\ 
\textbf{W}K_{21}      & \cdots &  \textbf{W}K_{2p} \\ 
\vdots          &                 \ddots    &   \vdots \\ 
 \textbf{W}K_{q1} & \cdots &  \textbf{W}K_{qp}
\end{pmatrix}$$
where $\bar{\textbf{W}} = reshape(\textbf{WK},[Lq,p])$. Finally, the PTC can be computed as 
$$
(F \star \textbf{K})= \left(\sum_{axis = 3} reshape(\textbf{Z}\bar{\textbf{W}},[p,n,q])\right)^T$$
Using the above sparse matrix operations, the computation complexity of the proposed PTC is the same scale as the standard convolution in Euclidean domains.  

{\small
\bibliographystyle{ieee}
\bibliography{PTCbib}
}

\end{document}